\documentclass[10pt]{article}
\usepackage[preprint]{tmlr}

\usepackage[T1]{fontenc}
\usepackage{amsmath,amssymb,amsfonts}
\usepackage{booktabs}
\usepackage{graphicx}
\usepackage{multirow}
\usepackage{xcolor}
\usepackage[colorlinks=true,linkcolor=blue!50!black,citecolor=blue!50!black,urlcolor=blue!50!black]{hyperref}
\usepackage{url}
\usepackage{algorithm}
\usepackage[noend]{algpseudocode}
\usepackage{caption}

\newcommand{\equalnote}{{\let\thefootnote\relax\footnotetext{$^{*}$Equal contribution.}}}
\newcommand{\acknowledgementsection}{\section*{Acknowledgements}
Generative AI tools were used to assist with language refinement and code development. The authors verified all results and take full responsibility for the content of this paper.
}
\newcommand{\method}{TripletAD}
\newcommand{\Ltrip}{\mathcal{L}_{\mathrm{trip}}}
\newcommand{\Lpce}{\mathcal{L}_{\mathrm{PCE}}}
\newcommand{\Lrc}{\mathcal{L}_{\mathrm{RC}}}

\title{\centering TripletAD: Weakly Supervised Anomaly Detection in\\Embedding Space with Rare Anomalies and Scarce Labels}

\author{\parbox{\linewidth}{\centering
      {\normalsize\bf Sumit Wadhwa$^{*}$ \qquad Souvik Nath$^{*}$}\\[2pt]
      {\small\normalfont Dell Technologies}\\[1pt]
      {\small\normalfont\ttfamily sumit.wadhwa@dell.com}}}

\def\month{MM}
\def\year{YYYY}
\def\openreview{\url{https://openreview.net/forum?id=XXXX}}

\begin{document}

\maketitle
\equalnote

\begin{abstract}
As AI and agentic systems automate more processes with less human review, data is increasingly represented, and passed between systems, as embeddings produced by pre-trained models. Detecting rare deviations in these embeddings is becoming a critical guardrail, yet anomaly detection methods were largely developed for hand-built tabular features, where gradient-boosted trees remain the method of choice. We propose \method{}, a weakly supervised detector for embedding spaces in which anomalies are rare. \method{} learns a triplet-margin embedding in which every triplet contains an anomaly, so the few anomalies are reused across the entire training set instead of being diluted by it, and calibrates a linear scoring head with a partial cross-entropy on anomalies and a relative-contrast hinge that separates the mean scores of anomalous and normal samples. Across 28 embedding datasets spanning text, images, speech and industrial inspection, \method{} is consistently among the best-performing methods, matching or exceeding established state-of-the-art weakly supervised detectors (DevNet, PReNet and DeepSAD), supervised contrastive learning and gradient-boosted trees, and it outperforms them when anomalies are rare. On naturally rare speech data it is 43\% above the strongest baseline and more than twice the established detectors; at a 1\% anomaly rate it has the highest mean AUCPR of seven methods, 20\% above DevNet; and pooled over 74 rare-anomaly settings it has the best average rank, with statistically significant gains over every baseline. In a case study on AI-agent traces, it locates the failing step more often than all six baselines, with accuracy on par with or above that reported for frontier LLM judges, at 0.17\,ms per step. The method needs no per-dataset tuning and is insensitive to its loss weights and training length.
\end{abstract}

\section{Introduction}
\label{sec:intro}

As AI and agentic systems automate more processes with less human review, data is increasingly represented as embeddings. Text, images, audio and logs are encoded by pre-trained models, and outputs are passed between models and agents in embedding form. With fewer people in the loop, reviewing these outcomes for deviations becomes one of the few available guardrails, and the same need runs through fraud review, defect inspection, content moderation and security monitoring. In these settings the deviations that matter are \emph{rare}, often 1--5\% of cases or fewer. Detecting rare anomalies in embedding space is therefore a growing and widespread need.

Historically, anomaly detection methods were developed largely for hand-built tabular features, where gradient-boosted trees such as CatBoost \citep{prokhorenkova2018catboost} remain the method to beat, ahead of deep detectors \citep{grinsztajn2022tree,han2022adbench}. On embedding inputs, the established state of the art for this weakly supervised setting is the family of deep detectors DevNet \citep{pang2019devnet}, PReNet \citep{pang2023prenet} and DeepSAD \citep{ruff2020deepsad}, designed for a few anomalies and a large pool of mostly normal data. When anomalies become very rare, two things work against them: each anomaly appears in only a small fraction of training batches unless it is explicitly oversampled, and the objective has few positive examples from which to shape the decision boundary. Rare anomalies in embedding space are therefore where improvement is most needed.

We take a metric-learning view. \method{} trains an encoder with a triplet-margin loss \citep{schroff2015facenet,hoffer2015triplet} over triplets mined so that \emph{every} triplet contains an anomaly: an anomalous anchor is paired with an anomalous positive and a normal negative, and a normal anchor with a normal positive and an anomalous negative. A linear head on the learned embedding is then calibrated with two soft terms: a partial cross-entropy that only rewards high scores on anomalies, and a relative-contrast hinge that pushes the mean anomaly score of anomalous samples above that of normal samples.

Our main finding is that \method{} is one of the strongest methods on embedding inputs, and that it outperforms established state-of-the-art detectors when anomalies are rare. On the naturally rare ADBench Speech dataset (1.65\% anomalies), it is 43\% above the strongest baseline and more than twice DevNet, PReNet and DeepSAD. Because most public embedding benchmarks ship with higher anomaly rates, we also lower the anomaly rate of eight text and image benchmarks to 2\% and 1\%: there \method{} has the highest mean AUCPR of the seven methods we compare, 20\% above DevNet and 53\% above CatBoost at 1\%. On four NLP-ADBench text datasets at their natural 3--4\% anomaly rates, it ties for the highest mean AUCPR and is best on three of the four, and on MVTec-AD \citep{bergmann2019mvtec} industrial inspection embeddings with few defects in training, it is again the best method. Pooled over all rare-anomaly settings, its gains over every baseline are statistically significant. Finally, the anomaly-anchored mining at the heart of \method{} transfers: added to a supervised contrastive learner, it lifts that method's accuracy substantially, while \method{} itself remains significantly better (Section~\ref{sec:transfer}). In a case study on failed AI-agent traces from Who\&When Pro \citep{liu2026whowhenpro}, \method{} locates the failing step more often than every baseline, with accuracy on par with or above that reported for frontier LLM judges at a fraction of the cost (Section~\ref{sec:agents}).

\paragraph{Contributions.}
\begin{itemize}
    \item \method{}, a simple weakly supervised detector that combines triplet-margin representation learning with two soft calibration penalties on a linear head (Section~\ref{sec:method}).
    \item An evaluation of rare-anomaly detection in embedding space, combining naturally rare data, benchmarks adjusted to 1--2\% anomalies, natural-rate text benchmarks and industrial inspection with few defects (Section~\ref{sec:setup}).
    \item An empirical study over 38 datasets and seven methods, with paired significance tests, showing that \method{} is among the best methods on embedding inputs and leads when anomalies are rare (Section~\ref{sec:results}).
    \item Evidence that the anomaly-anchored mining transfers: it substantially improves a supervised contrastive learner, while \method{}'s full design remains significantly better (Section~\ref{sec:transfer}).
    \item A case study on step-level failure attribution in AI-agent workflows, where \method{} is the most accurate of seven methods (Section~\ref{sec:agents}).
    \item Ablation and sensitivity analyses showing that the relative-contrast term is the key calibration component and that results are stable across loss weights and training lengths (Section~\ref{sec:ablation}).
\end{itemize}

\section{Related work}
\label{sec:related}

\paragraph{Weakly supervised anomaly detection.}
Surveys by \citet{chandola2009anomaly} and \citet{pang2021survey} cover the broader field. DevNet \citep{pang2019devnet} learns a scalar score with a deviation loss that pulls unlabelled samples towards a Gaussian reference and pushes labelled anomalies several standard deviations away, using batches balanced between labelled anomalies and unlabelled data. PReNet \citep{pang2023prenet} instead learns a pairwise relation score over pairs of anomalies and unlabelled samples. DeepSAD \citep{ruff2020deepsad} extends Deep SVDD \citep{ruff2018deepsvdd} by mapping unlabelled data near a centre and labelled anomalies away from it. ADBench \citep{han2022adbench} benchmarks these and many other detectors and supplies the datasets and reference implementations we use; NLP-ADBench \citep{li2024nlpadbench} extends the comparison to text. These methods treat the unlabelled pool as approximately normal, an assumption that is reasonable when anomalies are rare. Concurrently, ArcAD \citep{han2026arcad} calibrates reconstruction-based industrial image detectors with a few real defects; it operates on image patches inside a reconstruction model, whereas \method{} is a general detector for any embedding input.

\paragraph{Metric and contrastive learning.}
Siamese networks \citep{bromley1993signature,koch2015siamese} and triplet networks \citep{hoffer2015triplet,schroff2015facenet,hermans2017defense} learn embeddings in which distance reflects class membership, and supervised contrastive learning \citep{khosla2020supcon} generalises the idea to many positives per anchor. Siamese networks have also been applied to weakly supervised anomaly detection in industry, for example to flag grey-market transactions \citep{wadhwa2026patent}. There, pairs of records are labelled similar or dissimilar, with unlabelled records treated as normal, and a new record is classified by its pairwise similarity to a reference set of labelled anomalies. \method{} differs in three ways aimed at detection accuracy. First, each training triplet compares an anchor against both an anomaly and a normal sample, so the embedding learns a relative ordering rather than isolated pairwise similarities. Second, the few anomalies are reused across the whole training set, because every triplet contains one. Third, a scoring head is calibrated directly for ranking anomalies above normal samples, so the model outputs an anomaly score rather than a similarity to be thresholded. As a side benefit, scoring requires a single forward pass and no reference set. In anomaly detection, triplet or contrastive objectives have otherwise mostly been used in self-supervised form. TriAD \citep{sun2023triad} uses tri-domain contrastive learning for time-series anomalies; TripleAD \citep{xiao2025tripleAD} distils three channels for graph anomalies; and MulSen-TripleAD \citep{li2024mulsen} fuses three sensor modalities for object inspection. Despite the similar names, none of these use labelled anomalies inside the triplet or calibrate a score head as we do. \method{} uses the triplet loss as a supervised representation learner in which the scarce anomalies appear in every triplet.

\paragraph{Trees versus deep models on tabular data.}
\citet{grinsztajn2022tree} and \citet{gorishniy2021revisiting} show that tree ensembles remain strong on tabular data with heterogeneous, individually informative features. Our results on raw tabular anomaly datasets agree. On embedding inputs, produced by pre-trained models, our results point the other way.

\section{Method}
\label{sec:method}

\subsection{Setting}
We observe a training set $\mathcal{D} = \{(x_i, y_i)\}_{i=1}^{N}$ with $x_i \in \mathbb{R}^D$ and $y_i \in \{0, 1\}$, where $y_i = 1$ marks an anomaly and $y_i = 0$ a normal sample. Anomalies are few ($|\mathcal{A}| \ll N$, with $\mathcal{A} = \{i : y_i = 1\}$). The normal data may contain a few undetected anomalies; because anomalies are rare, we assume they have little effect. The goal is a score $s(x) \in [0, 1]$ that ranks anomalies above normal samples at test time.

\subsection{Architecture}
\method{} has an encoder $f_\theta : \mathbb{R}^D \to \mathbb{R}^d$ and a linear scoring head:
\begin{equation}
    z = f_\theta(x), \qquad s(x) = \sigma\!\left(w^\top z + b\right),
    \label{eq:score}
\end{equation}
where $\sigma$ is the logistic function. In all experiments $f_\theta$ is a multilayer perceptron with hidden widths 128 and 64, batch normalisation \citep{ioffe2015batchnorm} and ReLU activations, and an embedding size of $d = 32$. When the input is itself a pre-trained embedding (BERT, ResNet-18 or i-vector features), $f_\theta$ learns a task-specific 32-dimensional embedding on top of it. The pre-trained feature extractor is not fine-tuned.

\subsection{Anomaly-anchored triplet mining}
Each epoch uses every training sample once as an anchor $x_a$. For each anchor we draw a positive $x_p$ from the same class (excluding the anchor itself) and a negative $x_n$ from the opposite class, both uniformly at random:
\begin{itemize}
    \item anomalous anchor ($y_a = 1$): $x_p \in \mathcal{A}$, $x_n \notin \mathcal{A}$;
    \item normal anchor ($y_a = 0$): $x_p \notin \mathcal{A}$, $x_n \in \mathcal{A}$.
\end{itemize}
By construction, every triplet contains at least one anomaly. With $N$ samples and $|\mathcal{A}|$ anomalies, each anomaly appears in roughly $N / |\mathcal{A}|$ triplets per epoch, so a dataset with 1\% anomalies reuses each label about a hundred times per epoch without explicit oversampling or batch balancing. Mining is $O(N)$ per epoch.

\subsection{Objective}
For a batch of triplets, \method{} minimises
\begin{equation}
    J = \Ltrip + \lambda_1 \Lpce + \lambda_2 \Lrc .
    \label{eq:objective}
\end{equation}

\paragraph{Triplet loss.}
With $z_\star = f_\theta(x_\star)$ and Euclidean distance,
\begin{equation}
    \Ltrip = \frac{1}{B}\sum_{t=1}^{B} \max\!\left(0,\; \lVert z_a^{(t)} - z_p^{(t)} \rVert_2 - \lVert z_a^{(t)} - z_n^{(t)} \rVert_2 + m\right),
\end{equation}
with margin $m = 1$. The term clusters anomalies with anomalies and normals with normals in embedding space.

\paragraph{Partial cross-entropy (PCE).}
The head is rewarded for assigning high scores to anomalous anchors:
\begin{equation}
    \Lpce = -\frac{1}{B}\sum_{t=1}^{B} \mathbb{1}\!\left[y_a^{(t)} = 1\right] \log s\!\left(x_a^{(t)}\right).
\end{equation}
The loss is ``partial'' because normal anchors contribute nothing. In deployment the normal pool may still hide anomalies, so the head is never pushed to drive any particular normal sample's score to zero. The average runs over all $B$ triplets, so the term's weight scales with the anomaly rate. We found that averaging over anomalous anchors only was never better and was less stable (Appendix~\ref{app:tuning}).

\paragraph{Relative contrast (RC).}
Let $S_+$ and $S_-$ be the anomalous and normal members of the positives and negatives in the batch, $\{x_p^{(t)}, x_n^{(t)}\}_{t=1}^{B}$. Under our mining rule each triplet contributes exactly one sample to each set. RC is a hinge on the gap between their mean scores:
\begin{equation}
    \Lrc = \max\!\left(0,\; 1 - \Big(\tfrac{1}{|S_+|}\textstyle\sum_{x \in S_+} s(x) - \tfrac{1}{|S_-|}\sum_{x \in S_-} s(x)\Big)\right).
\end{equation}
RC is a soft, batch-level ranking constraint. It asks the head to separate the two groups on average rather than to classify each sample. It also supplies the negative-side signal that PCE deliberately omits.

\subsection{Training}
Training has two stages (Algorithm~\ref{alg:train}). Stage~1 is a short warm-up: the encoder is trained with $\Ltrip$ alone for 50 epochs, so the embedding already separates anomalies from normal samples before the scoring head is trained; the head receives no gradient in this stage. The warm-up gives a small gain on average, mainly when anomalies are rarest (Section~\ref{sec:ablation}). Stage~2 optimises the full objective~\eqref{eq:objective} for 100 epochs. Each batch stacks its anchors, positives and negatives into a single forward pass, so batch normalisation sees all three roles. We use Adam \citep{kingma2015adam} with learning rate $10^{-3}$, batches of 256 triplets, no weight decay, and the defaults $\lambda_1 = 1$ and $\lambda_2 = 0.25$ on every dataset. No hyperparameter is tuned per dataset. The anomaly score at test time is $s(x)$ from Eq.~\eqref{eq:score}.

\begin{algorithm}[t]
\caption{\method{} training}
\label{alg:train}
\begin{algorithmic}[1]
\Require training set $\mathcal{D}$, epochs $E_1 = 50$, $E_2 = 100$, weights $\lambda_1, \lambda_2$
\For{stage $\in \{1, 2\}$}
    \For{epoch $= 1, \dots, E_{\text{stage}}$}
        \State Mine one triplet per anchor $x_a \in \mathcal{D}$ (anomaly-anchored rule); shuffle into batches of 256
        \For{each batch}
            \State $J \gets \Ltrip$ \textbf{if} stage $= 1$ \textbf{else} $\Ltrip + \lambda_1 \Lpce + \lambda_2 \Lrc$
            \State Update $\theta$ (and $w, b$ in stage 2) with Adam
        \EndFor
    \EndFor
\EndFor
\State \Return $s(x) = \sigma(w^\top f_\theta(x) + b)$
\end{algorithmic}
\end{algorithm}

\section{Experimental setup}
\label{sec:setup}

\paragraph{Datasets.}
Table~\ref{tab:datasets} lists the datasets. Our focus is 28 datasets whose inputs are pre-trained embeddings. From ADBench \citep{han2022adbench}: Speech, whose inputs are 400-dimensional i-vectors \citep{rayana2016odds} and whose anomalies are naturally rare; five text corpora encoded with BERT \citep{devlin2019bert} (20News, AGNews, Amazon, IMDB, Yelp) and three image datasets encoded with ResNet-18 \citep{he2016resnet} (CIFAR-10, FashionMNIST, SVHN), using ADBench's first class split (class 0 is the anomaly class); and the 15 MVTec-AD industrial inspection categories \citep{bergmann2019mvtec}, encoded with ResNet-18. From NLP-ADBench \citep{li2024nlpadbench}: four text datasets (BBC news, SMS spam, email spam and N24News) with their BERT embeddings. NLP-ADBench places all anomalies in its test files, so we pool its train and test files and apply our own split, which keeps each dataset's natural 3--4\% anomaly rate in both parts. For comparison, we also evaluate ten raw tabular datasets from ADBench.

\begin{table}[!htbp]
\centering
\caption{Datasets. ``Best 1-feature AUROC'' is the highest AUROC obtained by ranking on a single input coordinate (either sign).}
\label{tab:datasets}
\small
\begin{tabular}{lrrrrc}
\toprule
Dataset & $N$ & $D$ & Anomalies & Rate (\%) & Best 1-feature AUROC \\
\midrule
\multicolumn{6}{l}{\emph{Embedding (text: BERT; image: ResNet-18)}} \\
\quad 20News & 3,090 & 768 & 154 & 4.98 & 0.708 \\
\quad AGNews & 10,000 & 768 & 500 & 5.00 & 0.730 \\
\quad Amazon & 10,000 & 768 & 500 & 5.00 & 0.713 \\
\quad IMDB & 10,000 & 768 & 500 & 5.00 & 0.698 \\
\quad Yelp & 10,000 & 768 & 500 & 5.00 & 0.753 \\
\quad CIFAR-10 & 5,263 & 512 & 263 & 5.00 & 0.782 \\
\quad FashionMNIST & 6,315 & 512 & 315 & 4.99 & 0.779 \\
\quad SVHN & 5,208 & 512 & 260 & 4.99 & 0.723 \\
\multicolumn{6}{l}{\emph{Embedding (speech: i-vectors)}} \\
\quad Speech & 3,686 & 400 & 61 & 1.65 & 0.674 \\
\multicolumn{6}{l}{\emph{Embedding (NLP-ADBench text: BERT)}} \\
\quad BBC news & 1,785 & 768 & 62 & 3.47 & 0.894 \\
\quad SMS spam & 4,672 & 768 & 154 & 3.30 & 0.950 \\
\quad Email spam & 3,578 & 768 & 146 & 4.08 & 0.875 \\
\quad N24News & 59,822 & 768 & 1,828 & 3.06 & 0.854 \\
\multicolumn{6}{l}{\emph{Raw tabular}} \\
\quad Backdoor & 95,329 & 196 & 2,329 & 2.44 & 0.887 \\
\quad Cover & 286,048 & 10 & 2,747 & 0.96 & 0.999 \\
\quad Fraud & 284,807 & 29 & 492 & 0.17 & 0.933 \\
\quad HTTP & 567,498 & 3 & 2,211 & 0.39 & 1.000 \\
\quad Waveform & 3,443 & 21 & 100 & 2.90 & 0.873 \\
\quad Yeast & 1,484 & 8 & 507 & 34.16 & 0.592 \\
\quad ALOI & 49,534 & 27 & 1,508 & 3.04 & 0.569 \\
\quad Campaign & 41,188 & 62 & 4,640 & 11.27 & 0.818 \\
\quad CelebA & 202,599 & 39 & 4,547 & 2.24 & 0.793 \\
\quad Census & 299,285 & 500 & 18,568 & 6.20 & 0.787 \\
\multicolumn{6}{l}{\emph{Embedding (MVTec-AD, ResNet-18), 15 categories}} \\
\quad range & 102--501 & 512 & 30--141 & 12.8--32.5 & 0.601--0.996 \\
\bottomrule
\end{tabular}

\end{table}

\paragraph{Splits and preprocessing.}
For each of five seeds we make a stratified 70/30 train/test split and fit min--max scaling on the training part only. Every method sees exactly the same split, scaling and labels for a given seed. We report AUCPR (average precision), which is more informative than AUROC under heavy class imbalance \citep{davis2006prauc}, averaged over the five seeds.

\paragraph{Rare anomalies.}
Most public embedding benchmarks ship with anomaly rates of 5\% or more. To study the rare regime, we lower the anomaly rate of the eight ADBench text and image datasets by subsampling their anomalies, keeping all normal samples, so that anomalies make up 2\% or 1\% of the data. The subsample is redrawn for each of the five seeds, so the reported variance includes which anomalies were kept. At 1\% the training sets contain 21 (20News) to 67 (AGNews, Amazon, IMDB, Yelp) anomalies, comparable to Speech (43). A single fixed subsample, reused for all seeds, gives the same conclusions (Table~\ref{tab:rarityfixed}).

\paragraph{Fewer anomalies in training.}
To vary how many anomalies the model sees, we keep a random fraction $\rho$ of the training anomalies and remove the rest from the training set. The test set is never modified. We skip a (dataset, $\rho$) cell when fewer than ten training anomalies would remain. On MVTec-AD, keeping a quarter of the defects lowers the share of defects in training to 4--10\%. Tables report each setting by the number of training anomalies and their share of the training data.

\paragraph{Baselines.}
We compare against two tree ensembles, three deep weakly supervised detectors and a supervised contrastive learner, all trained on identical inputs:
\begin{itemize}
    \item random forest \citep{breiman2001random}: scikit-learn, 100 trees;
    \item CatBoost \citep{prokhorenkova2018catboost}: library defaults;
    \item DevNet \citep{pang2019devnet}: our re-implementation following the original paper, with one hidden layer of 20 units, deviation loss with margin 5, and balanced batches of 512;
    \item PReNet \citep{pang2023prenet} and DeepSAD \citep{ruff2020deepsad}: ADBench's reference implementations at their default hyperparameters;
    \item SupCon \citep{khosla2020supcon}, the closest relative of our triplet objective: supervised contrastive learning with the same encoder and training budget as \method{} (50 contrastive epochs on shuffled batches of 256, temperature 0.1), followed by a linear head trained with class-weighted cross-entropy on the frozen embedding.
\end{itemize}
No method, including ours, is tuned per dataset. ADBench's DeepSAD fails on a few very small MVTec training sets (its data loader yields zero batches). Those cells are omitted for DeepSAD and marked in the tables.

\paragraph{Statistics.}
Following \citet{demsar2006statistical}, the unit of analysis is a (dataset, setting) cell, with AUCPR averaged over seeds. We compare \method{} with each baseline using a two-sided Wilcoxon signed-rank test over paired cells and correct the resulting $p$-values with the Holm procedure \citep{holm1979simple}. We also report average ranks and a Friedman test over cells where all methods have a result. Every test anomaly is unseen during training, so all reported scores measure generalisation to new anomalies.

\paragraph{Compute.}
All runs use a CPU with one or two threads per process; no GPU is needed.

\section{Results: rare anomalies in embedding space}
\label{sec:results}

We start with naturally rare data, then turn to benchmarks whose anomaly rate we lowered to the rare range, natural-rate text benchmarks, and industrial inspection with few defects in training. All seven methods, including the tree ensembles, are compared in every setting.

\subsection{Naturally rare anomalies: Speech}

Speech is the dataset closest to the target setting: anomalies are naturally rare (1.65\%), each training split holds only 43 of them, and the i-vector coordinates are individually weak. Here \method{}'s advantage is largest (Table~\ref{tab:speech}). Trained on all 43 anomalies, it reaches 0.394 AUCPR, 43\% above the strongest baseline (SupCon, 0.275), twice CatBoost (0.196) and more than twice DevNet (0.178), PReNet (0.165) and DeepSAD (0.073). The lead holds with even fewer anomalies in training: 0.317 with 32 and 0.252 with 21, against at most 0.249 and 0.195 for any baseline. With only ten training anomalies, all methods are close, with SupCon, PReNet and DevNet slightly ahead.

\begin{table}[!htbp]
\centering
\caption{Speech (1.65\% anomalies): AUCPR as the number of training anomalies varies, with their share of the training data in brackets.}
\label{tab:speech}
\small
\begin{tabular}{lccccccc}
\toprule
Training anomalies (share) & TripletAD & SupCon & DeepSAD & DevNet & PReNet & CatBoost & RF \\
\midrule
43 (1.67\%) & \textbf{0.394} & 0.275 & 0.073 & 0.178 & 0.165 & 0.196 & 0.070 \\
32 (1.25\%) & \textbf{0.317} & 0.249 & 0.060 & 0.180 & 0.181 & 0.152 & 0.068 \\
21 (0.82\%) & \textbf{0.252} & 0.195 & 0.050 & 0.157 & 0.163 & 0.113 & 0.071 \\
10 (0.39\%) & 0.093 & \textbf{0.112} & 0.032 & 0.106 & 0.109 & 0.053 & 0.036 \\
\bottomrule
\end{tabular}

\end{table}

\subsection{Benchmarks lowered to 1\% and 2\% anomalies}

\paragraph{1\% anomalies.}
With anomalies at 1\% of the data (Table~\ref{tab:rarity}), \method{} has the highest mean AUCPR of the seven methods (0.244), ahead of SupCon (0.215), DeepSAD (0.210), PReNet (0.207), DevNet (0.204), CatBoost (0.159) and random forest (0.118). That is 20\% above DevNet and 53\% above CatBoost. It is the best method, or tied for best, on five of the eight datasets, and its average rank is 2.12, against 3.25 for the next best. The text datasets are where it is strongest: it matches or beats DevNet and PReNet on all five, and on AGNews it nearly doubles them (0.325 against 0.175 and 0.171).

\paragraph{2\% anomalies.}
At 2\%, \method{} again has the highest mean AUCPR (0.346, against 0.333 for DevNet, 0.329 for SupCon, 0.326 for DeepSAD and 0.289 for CatBoost) and the best average rank (2.25).

\begin{figure}[!htbp]
\centering
\includegraphics[width=\linewidth]{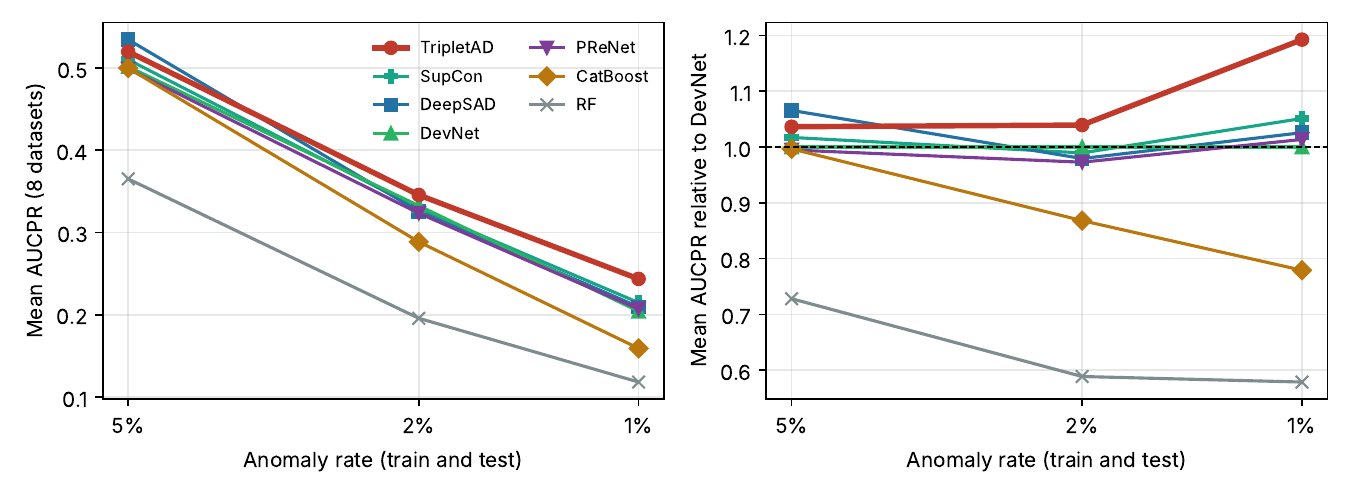}
\caption{Eight text and image embedding datasets as the anomaly rate falls from 5\% to 1\%. \textbf{Left:} mean AUCPR over datasets and five seeds. \textbf{Right:} the same, divided by DevNet's score. \method{}'s lead grows to 20\% over DevNet at 1\%, while the tree ensembles fall further behind.}
\label{fig:rarity}
\end{figure}

\paragraph{The advantage grows as anomalies become rarer.}
Figure~\ref{fig:rarity} follows the same eight datasets from their original 5\% anomaly rate down to 1\%. Every method loses AUCPR as anomalies thin out, but the baselines lose it faster: \method{}'s lead over DevNet grows from about 4\% to 20\%, and the tree ensembles fall further behind. Across the 16 cells at 1\% and 2\%, it beats random forest on all 16, CatBoost on 15, DeepSAD on 13, SupCon on 12, PReNet on 11 and DevNet on 10; the pooled analysis of Section~\ref{sec:stats} confirms these gains statistically.

\begin{table}[!htbp]
\centering
\caption{Text and image benchmarks with the anomaly rate lowered to 1\% and 2\%: AUCPR (mean of five seeds; the anomaly subsample is redrawn for each seed). Best per row in bold.}
\label{tab:rarity}
\small
\begin{tabular}{lccccccc}
\toprule
Dataset & TripletAD & SupCon & DeepSAD & DevNet & PReNet & CatBoost & RF \\
\midrule
\multicolumn{8}{l}{\emph{Anomaly rate 1\%}} \\
\quad 20News & \textbf{0.341} & 0.269 & 0.212 & \textbf{0.341} & 0.329 & 0.152 & 0.106 \\
\quad AGNews & \textbf{0.325} & 0.274 & 0.262 & 0.175 & 0.171 & 0.159 & 0.082 \\
\quad Amazon & \textbf{0.111} & 0.108 & 0.095 & 0.099 & 0.107 & 0.057 & 0.023 \\
\quad IMDB & \textbf{0.126} & 0.049 & 0.118 & 0.095 & 0.081 & 0.069 & 0.022 \\
\quad Yelp & 0.181 & 0.155 & \textbf{0.196} & 0.151 & 0.150 & 0.117 & 0.047 \\
\quad CIFAR-10 & 0.135 & 0.151 & 0.130 & 0.179 & \textbf{0.186} & 0.142 & 0.116 \\
\quad FashionMNIST & \textbf{0.575} & 0.539 & 0.573 & 0.411 & 0.431 & 0.490 & 0.491 \\
\quad SVHN & 0.154 & 0.174 & 0.092 & 0.183 & \textbf{0.203} & 0.089 & 0.061 \\
\quad \emph{Average} & \textbf{0.244} & 0.215 & 0.210 & 0.204 & 0.207 & 0.159 & 0.118 \\
\midrule
\multicolumn{8}{l}{\emph{Anomaly rate 2\%}} \\
\quad 20News & 0.441 & \textbf{0.476} & 0.396 & 0.463 & 0.447 & 0.367 & 0.250 \\
\quad AGNews & 0.332 & 0.340 & \textbf{0.392} & 0.304 & 0.300 & 0.293 & 0.169 \\
\quad Amazon & \textbf{0.240} & 0.161 & 0.169 & 0.158 & 0.152 & 0.142 & 0.048 \\
\quad IMDB & \textbf{0.197} & 0.185 & 0.182 & 0.168 & 0.161 & 0.136 & 0.061 \\
\quad Yelp & 0.319 & 0.285 & \textbf{0.364} & 0.294 & 0.288 & 0.261 & 0.099 \\
\quad CIFAR-10 & 0.281 & 0.248 & 0.217 & \textbf{0.295} & 0.287 & 0.250 & 0.191 \\
\quad FashionMNIST & \textbf{0.660} & 0.653 & 0.621 & 0.603 & 0.600 & 0.612 & 0.575 \\
\quad SVHN & 0.295 & 0.284 & 0.266 & \textbf{0.376} & 0.353 & 0.247 & 0.173 \\
\quad \emph{Average} & \textbf{0.346} & 0.329 & 0.326 & 0.333 & 0.324 & 0.289 & 0.196 \\
\bottomrule
\end{tabular}

\end{table}

\subsection{Natural anomaly rates on text benchmarks: NLP-ADBench}

To check the result on data whose anomaly rate we did not adjust, we evaluate four NLP-ADBench datasets at their natural 3--4\% anomaly rates (Table~\ref{tab:nlp}). \method{} ties with SupCon for the highest mean AUCPR (0.936) and is the best method on three of the four: BBC news (0.954), SMS spam (0.949) and email spam (0.963). It outperforms DevNet and PReNet on all four datasets, by 0.02 to 0.10 AUCPR, and random forest on all four. On N24News, SupCon (0.924), DeepSAD (0.920) and CatBoost (0.897) are ahead of \method{} (0.878). \method{} is also among the most stable methods across seeds, with the smallest spread on SMS spam and email spam.

\begin{table}[!htbp]
\centering
\caption{NLP-ADBench text datasets (BERT embeddings) at their natural anomaly rates (3.1--4.1\%): AUCPR (mean of five seeds). Best per row in bold.}
\label{tab:nlp}
\small
\begin{tabular}{lccccccc}
\toprule
Dataset & TripletAD & SupCon & DeepSAD & DevNet & PReNet & CatBoost & RF \\
\midrule
BBC news & \textbf{0.954} & 0.937 & 0.938 & 0.908 & 0.906 & 0.849 & 0.758 \\
SMS spam & \textbf{0.949} & 0.938 & 0.936 & 0.926 & 0.926 & 0.912 & 0.880 \\
Email spam & \textbf{0.963} & 0.947 & 0.933 & 0.891 & 0.860 & 0.861 & 0.773 \\
N24News & 0.878 & \textbf{0.924} & 0.920 & 0.810 & 0.798 & 0.897 & 0.784 \\
\midrule
Average & \textbf{0.936} & \textbf{0.936} & 0.932 & 0.884 & 0.872 & 0.880 & 0.799 \\
\bottomrule
\end{tabular}

\end{table}

\subsection{Few defects in training: MVTec-AD industrial inspection}

On the 15 MVTec-AD categories (Table~\ref{tab:mvtec}), \method{} is the best method when few defects are available for training. With a quarter of the defects used, which leaves 4--10\% defects in training, its mean AUCPR is 0.704, ahead of CatBoost (0.698), PReNet (0.683), SupCon (0.679), DevNet (0.671) and DeepSAD (0.625). With half used it again leads (0.734). Over the 27 category cells at 25\% and 50\% where all seven methods ran, it has the best average rank (2.81). With most or all defects used, the methods converge: CatBoost, SupCon, DevNet and \method{} are within 0.013 of each other. Per-category results are in Table~\ref{tab:mvtecfull}.

\begin{table}[!htbp]
\centering
\caption{MVTec-AD (ResNet-18 embeddings): mean AUCPR over categories as the number of training defects varies (range over categories, with their share of the training data in brackets). Each row keeps the same fraction of every category's defects (a quarter, half, three quarters, all). ``Categories'': number of the 15 categories with at least ten training defects in that row, over which the mean is taken. ``TripletAD best in'': number of those categories in which \method{} has the highest AUCPR of the seven methods. $^\dagger$DeepSAD mean over the categories where it ran.}
\label{tab:mvtec}
\small
\setlength{\tabcolsep}{4pt}
\begin{tabular}{lccccccccc}
\toprule
\shortstack{Training defects\\(share)} & Categories & TripletAD & SupCon & DeepSAD & DevNet & PReNet & CatBoost & RF & \shortstack{TripletAD\\best in} \\
\midrule
10--24 (4--10\%) & 13 & \textbf{0.704} & 0.679 & 0.625 & 0.671 & 0.683 & 0.698 & 0.691 & 5 \\
10--49 (7--19\%) & 15 & \textbf{0.734} & 0.721 & 0.644$^\dagger$ & 0.714 & 0.710 & 0.732 & 0.719 & 6 \\
15--73 (10--26\%) & 15 & 0.742 & 0.742 & 0.665$^\dagger$ & 0.735 & 0.726 & \textbf{0.748} & 0.733 & 5 \\
21--98 (13--32\%) & 15 & 0.745 & 0.755 & 0.676$^\dagger$ & 0.749 & 0.737 & \textbf{0.758} & 0.742 & 3 \\
\bottomrule
\end{tabular}

\end{table}

\subsection{Statistical summary}
\label{sec:stats}
Table~\ref{tab:stats} pools every rare-anomaly embedding setting: Speech, the 1\% and 2\% benchmarks, the NLP-ADBench datasets, MVTec-AD with a quarter or half of the defects used, and the text and image benchmarks with fewer training anomalies (Appendix~\ref{app:full}), for 74 cells in total. \method{} has the best average rank over the 72 cells where every method ran (2.38, against 3.32 for DevNet and 3.56 for PReNet; Friedman $p = 2 \times 10^{-20}$). It also outperforms every baseline individually after Holm correction, by 0.023--0.038 AUCPR on average over DevNet, PReNet and SupCon, and by 0.05--0.10 over DeepSAD and the trees.

\begin{table}[!htbp]
\centering
\caption{Pooled comparison over 74 rare-anomaly embedding cells (AUCPR averaged over five seeds per cell). W/L: cells where \method{} is better/worse. $p$: two-sided Wilcoxon signed-rank, Holm-corrected over the six baselines. A few cells are missing for DeepSAD; one DevNet and one PReNet cell are tied.}
\label{tab:stats}
\small
\begin{tabular}{lcccc}
\toprule
\method{} vs. & Cells & W / L & Mean AUCPR gain & Holm $p$ \\
\midrule
DeepSAD  & 72 & 65 / 7  & $+0.068$ & $2 \times 10^{-10}$ \\
Random forest & 73 & 61 / 12 & $+0.102$ & $8 \times 10^{-10}$ \\
SupCon   & 73 & 58 / 15 & $+0.038$ & $2 \times 10^{-7}$ \\
CatBoost & 73 & 57 / 16 & $+0.054$ & $2 \times 10^{-7}$ \\
PReNet   & 73 & 48 / 24 & $+0.025$ & $0.003$ \\
DevNet   & 73 & 48 / 24 & $+0.023$ & $0.006$ \\
\bottomrule
\end{tabular}

\vspace{3pt}
{\footnotesize Average rank over 72 complete cells (1 = best): \method{} 2.38, DevNet 3.32, PReNet 3.56, SupCon 3.92, CatBoost 4.43, DeepSAD 4.74, RF 5.67.}
\end{table}

\subsection{Anomaly-anchored mining transfers to other methods}
\label{sec:transfer}
The ingredient behind \method{} is a sampling principle: every training triplet contains an anomaly, so the few anomalies are reused across the whole training set. To test whether this principle helps beyond our objective, we train SupCon on batches built from \method{}'s anomaly-anchored triplets (``+ TripletAD mining''). For reference, we also train it on batches in which half the samples are anomalies (``+ balanced batches''), the sampling used by DevNet. Table~\ref{tab:transfer} shows that anomaly-focused sampling lifts SupCon substantially: on Speech from 0.275 to 0.321, and at 1\% anomalies from 0.215 to 0.244. \method{}'s full design is still the strongest. Pooled over the same 74 cells, it outperforms SupCon with \method{} mining on 50 cells and loses on 22 (mean gain $+0.015$, Holm $p < 0.001$), and SupCon with balanced batches on 51 and loses on 22 ($+0.011$, $p = 0.007$). Among the four variants it has the best average rank (1.82, against 2.33 and 2.49 for the two enhanced SupCon variants and 3.36 for SupCon as published). Its lead is largest where anomalies are naturally rare (Speech) or few anomalies are available for training (MVTec-AD), while on the 1\% and 2\% benchmarks and NLP-ADBench the enhanced variants are level with or slightly ahead of \method{}.

\begin{table}[!htbp]
\centering
\caption{SupCon as published, SupCon with anomaly-focused sampling, and \method{}: mean AUCPR over five seeds. Best per row in bold.}
\label{tab:transfer}
\small
\begin{tabular}{lcccc}
\toprule
Setting & SupCon & + TripletAD mining & + balanced batches & TripletAD \\
\midrule
Speech (1.65\%) & 0.275 & 0.321 & 0.323 & \textbf{0.394} \\
Text \& image, 1\% & 0.215 & 0.244 & \textbf{0.251} & 0.244 \\
Text \& image, 2\% & 0.329 & 0.343 & \textbf{0.353} & 0.346 \\
NLP-ADBench (3--4\%) & 0.936 & \textbf{0.945} & \textbf{0.945} & 0.936 \\
MVTec-AD, 25\% of defects & 0.679 & 0.688 & 0.683 & \textbf{0.704} \\
MVTec-AD, 50\% of defects & 0.721 & 0.729 & 0.726 & \textbf{0.734} \\
\bottomrule
\end{tabular}

\end{table}

\paragraph{Cost.}
\method{} is also fast to train and to apply. On the eight text and image datasets with all anomalies used, one fit plus scoring takes 28\,s on a single CPU thread on average, against 33\,s for PReNet, 39\,s for DeepSAD, 41\,s for DevNet and 130\,s for CatBoost (random forest: 23\,s; SupCon: 4\,s, with fewer optimisation steps per epoch). Scoring a new sample is a single forward pass, taking 0.17\,ms on one CPU thread.

\subsection{Case study: locating the failing step in agent traces}
\label{sec:agents}
To test \method{} beyond standard anomaly-detection benchmarks, we apply it to failure attribution in AI-agent workflows. Who\&When Pro \citep{liu2026whowhenpro} injects one error at a known step into otherwise correct agent trajectories and asks which step is the first decisive error. We use the 1,210 traces whose files are present in its public release: 311 single-agent ALFWorld household tasks \citep{shridhar2021alfworld}, which are text, and 899 multi-agent LVBench video question-answering traces \citep{wang2024lvbench}, whose messages and tool calls are text. Each step (an agent's message or action together with the result it receives) is encoded with the all-MiniLM-L6-v2 sentence encoder \citep{reimers2019sbert} (384 dimensions); the label marking the injected step is never part of the encoded text. Because the two benchmarks come from different agents, each method is trained and tested separately on each, as one detector per agent type would be deployed. The traces of each benchmark are split 70/30, so no trace contributes to both training and test. Every method is trained on the steps of the training traces up to and including the injected step, with the injected step as the anomaly; steps after it are left out because the error can propagate into them. On each of the 363 test traces (93 ALFWorld, 270 LVBench), every step is scored, including those after the error, and the highest-scoring step is the prediction. This is the benchmark's step-level accuracy. ALFWorld traces have 30 steps on average and LVBench traces two to three, so picking a step at random is right 3.3\% and 46.7\% of the time.

Table~\ref{tab:agent} shows that \method{} locates the failing step in 68.8\% of test traces, the highest of the seven methods, ahead of DeepSAD (65.3\%) and SupCon (64.8\%). It is the most accurate method overall and on LVBench (65.0\%, ahead of DeepSAD at 62.7\% and the random forest at 62.1\%), and level with SupCon on ALFWorld (79.6\% against 79.4\%), well ahead of the remaining methods there (41--73\%). For context, \citet{liu2026whowhenpro} report step-level accuracies of up to 73.9\% for zero-shot frontier LLM judges on text traces and up to 64.4\% on LVBench. \method{}'s accuracy is on par with or above these, while scoring a step takes 0.17\,ms on a single CPU thread with precomputed step embeddings. Training one detector per agent type matters: a single detector trained on both benchmarks together is less accurate on ALFWorld for every method (73.4\% for \method{}), although \method{} remains the most accurate overall (68.5\%).

The comparison with LLM judges is not like for like, and we report it only for context. The judges are zero-shot and read the whole trace, whereas \method{} is trained on labelled failures from the same benchmarks; their text figure covers 16 text benchmarks of which ALFWorld is one, and they evaluate many more traces than the public subset we use. The case study also covers only step-level attribution: the benchmark additionally scores which agent failed and the type of error, which an anomaly score does not provide. Within these limits, the results indicate that rare-anomaly detection in embedding space is a practical and inexpensive tool for locating failures in agent workflows.

\begin{table}[!htbp]
\centering
\caption{Case study: step-level failure attribution on the public Who\&When Pro traces, with one detector per agent type. Percentage of test traces in which the highest-scoring step is the injected error step (mean $\pm$ std over five seeds; best per column in bold, with ties within 0.25 points). The bottom rows are reported by \citet{liu2026whowhenpro} for zero-shot LLM judges on the full benchmark and are not directly comparable. $^\dagger$Over 16 text benchmarks, of which ALFWorld is one.}
\label{tab:agent}
\small
\begin{tabular}{lccc}
\toprule
Method & All (363 traces) & ALFWorld (text) & LVBench (video) \\
\midrule
TripletAD & \textbf{68.8} $\pm$ 3.1 & \textbf{79.6} $\pm$ 9.0 & \textbf{65.0} $\pm$ 3.7 \\
DeepSAD & 65.3 $\pm$ 2.9 & 72.7 $\pm$ 5.4 & 62.7 $\pm$ 2.8 \\
SupCon & 64.8 $\pm$ 2.2 & \textbf{79.4} $\pm$ 4.9 & 59.8 $\pm$ 3.0 \\
PReNet & 61.0 $\pm$ 3.1 & 65.8 $\pm$ 5.2 & 59.4 $\pm$ 3.0 \\
DevNet & 60.6 $\pm$ 3.2 & 63.7 $\pm$ 5.0 & 59.6 $\pm$ 3.1 \\
CatBoost & 58.3 $\pm$ 3.7 & 54.2 $\pm$ 7.4 & 59.8 $\pm$ 3.3 \\
RF & 56.8 $\pm$ 3.5 & 41.5 $\pm$ 6.1 & 62.1 $\pm$ 3.9 \\
Random step & 35.6 & 3.3 & 46.7 \\
\midrule
\multicolumn{4}{l}{\emph{Reported in \citet{liu2026whowhenpro}: zero-shot LLM judges, full benchmark}} \\
Best text judge (Qwen3.5-122B) & & 73.9$^\dagger$ & \\
Best video judge (Claude Sonnet 4.6) & & & 64.4 \\
\bottomrule
\end{tabular}

\end{table}

\section{Ablation and sensitivity}
\label{sec:ablation}

\paragraph{Components.}
Table~\ref{tab:ablation} removes each component on six embedding datasets where \method{} performs strongly. The full objective has the highest mean AUCPR (0.693). Removing RC is the most damaging change: the mean falls to 0.665, with the largest drops on FashionMNIST (0.575 to 0.508), AGNews (0.325 to 0.289) and BBC news (0.954 to 0.925). Removing PCE or the triplet warm-up costs little on most datasets but clearly hurts on AGNews at 1\% anomalies (0.325 to 0.301 and 0.280). With neither calibration term, the head is never trained and the scores collapse (mean 0.186), which confirms that the embedding alone is not a detector. RC therefore carries most of the calibration, and PCE and the warm-up are small refinements that matter most when anomalies are rarest.

\begin{table}[!htbp]
\centering
\caption{Ablation on six embedding datasets: AUCPR (mean over five seeds), all training anomalies used. ``1\%'': anomaly rate lowered to 1\%.}
\label{tab:ablation}
\small
\begin{tabular}{lccccccc}
\toprule
Variant & Speech & BBC & Email & SMS & AGNews 1\% & F-MNIST 1\% & Mean \\
\midrule
Full objective & 0.394 & 0.954 & 0.963 & 0.949 & 0.325 & 0.575 & 0.693 \\
w/o PCE ($\lambda_1{=}0$) & 0.391 & 0.954 & 0.967 & 0.949 & 0.301 & 0.576 & 0.689 \\
w/o RC ($\lambda_2{=}0$) & 0.380 & 0.925 & 0.952 & 0.938 & 0.289 & 0.508 & 0.665 \\
Triplet only & 0.129 & 0.203 & 0.232 & 0.256 & 0.057 & 0.236 & 0.186 \\
w/o triplet warm-up & 0.380 & 0.957 & 0.965 & 0.951 & 0.280 & 0.591 & 0.687 \\
\bottomrule
\end{tabular}

\end{table}

\paragraph{Loss weights and training length.}
Figure~\ref{fig:robustness} in Appendix~\ref{app:tuning} (left) shows mean test AUCPR on nine embedding datasets (the eight text and image datasets plus Speech) over a grid of $\lambda_1 \in \{0.1, 0.5, 1\}$ and $\lambda_2 \in \{0.05, 0.1, 0.25, 0.5\}$, with three seeds each. All twelve settings fall between 0.477 and 0.495. The default (1, 0.25) scores 0.483, within 0.012 of the best setting (0.1, 0.25). We keep the default to avoid tuning on the evaluation datasets. Figure~\ref{fig:robustness} (right) shows test AUCPR as stage-2 training runs from 10 to 200 epochs on four embedding datasets (Speech, SMS spam, BBC news and AGNews at 1\% anomalies). The curves are flat beyond about 50 epochs; the small fluctuations on AGNews reflect its few held-out anomalies. \method{} needs no early stopping or validation-based tuning in our experiments.

\section{Scope and limitations}
\label{sec:limitations}
\label{sec:scope}

\begin{itemize}
    \item \textbf{More frequent anomalies.} The advantage of \method{} is specific to rarity. At the ADBench default of about 5\% anomalies, the methods are close: \method{} (0.520 mean AUCPR) is on par with SupCon, DevNet, PReNet and CatBoost, and DeepSAD is slightly ahead (0.535) (Tables~\ref{tab:highdimfull} and~\ref{tab:rarityfull}). With fewer anomalies used at that rate, \method{} is on par with DevNet and PReNet (Tables~\ref{tab:sparse} and~\ref{tab:sparsefull}), and on MVTec-AD with all defects used, the methods converge.
    \item \textbf{Very few training anomalies.} With 10--20 training anomalies, \method{} remains competitive (top three in 20 of 25 such settings, never last), but no method leads consistently. We do not evaluate fewer than 10 training anomalies.
    \item \textbf{Raw tabular data.} On raw tabular features, gradient-boosted trees remain the stronger choice (Table~\ref{tab:tabular}), consistent with prior work \citep{grinsztajn2022tree}; among deep detectors, \method{} still outperforms DevNet on seven of ten datasets.
    \item \textbf{Individual datasets.} At 1\% anomalies, PReNet and DevNet remain ahead on CIFAR-10 and SVHN, and SupCon and DeepSAD on N24News.
    \item \textbf{Adjusted anomaly rates.} The 1\% and 2\% datasets are made by subsampling anomalies from existing benchmarks, because few public embedding datasets are naturally that rare; Speech, the naturally rare case, shows the largest advantage. Their test sets contain few anomalies (9--29), so our conclusions rest on pooled, paired tests rather than individual cells.
    \item \textbf{Agent-trace case study.} The case study covers step-level attribution on the public traces of one benchmark, with each trace scored after it has finished. Attributing failures to agents and error types, and flagging them while an agent runs, are future work.
    \item \textbf{Frozen features and default settings.} \method{} learns on fixed pre-trained features, and no method is tuned per dataset. End-to-end training is future work; the loss-weight study (Appendix~\ref{app:tuning}, Figure~\ref{fig:robustness}) suggests tuning would gain little.
\end{itemize}

\section{Conclusion}
\label{sec:conclusion}

\method{} combines triplet-margin representation learning, anomaly-anchored mining and two soft calibration penalties into a weakly supervised detector with no per-dataset tuning. On embedding inputs it is consistently among the best-performing methods, and it outperforms established state-of-the-art weakly supervised detectors and tree ensembles when anomalies are rare: it is 43\% above the strongest baseline on naturally rare speech data, it is the strongest of the seven methods we compare at a 1\% anomaly rate, it is best on three of four natural-rate text benchmarks, and pooled over 74 rare-anomaly settings it ranks first, with significant gains over each baseline. In a case study on AI-agent traces, it is also the most accurate of the seven methods at locating the failing step.

These results matter most where anomalies are becoming rarer. In manufacturing, defect rates fall as processes mature, and our results show \method{}'s advantage growing as anomalies become rarer, down to the 1\% rate we tested. As AI and agentic systems automate more processes with less human review, and exchange their outputs as embeddings, detecting rare deviations in those outputs becomes an important guardrail. Our case study on agent traces is a first step in this direction; using \method{} to flag anomalous agent behaviour as it happens is a natural next step.

\paragraph{Reproducibility.}
Code, data-preparation scripts pinned to public dataset versions, and per-run results for every table and figure are available from the authors upon request. All datasets are public through ADBench, NLP-ADBench and Who\&When Pro.

\acknowledgementsection

\bibliography{refs}
\bibliographystyle{tmlr}

\appendix
\section{Additional studies}
\label{app:tuning}

\paragraph{PCE averaging.}
Averaging $\Lpce$ over anomalous anchors only, instead of over all triplets, was never better than the default in our preliminary experiments and was less stable. We keep averaging over all triplets.

\begin{figure}[h]
\centering
\includegraphics[width=\linewidth]{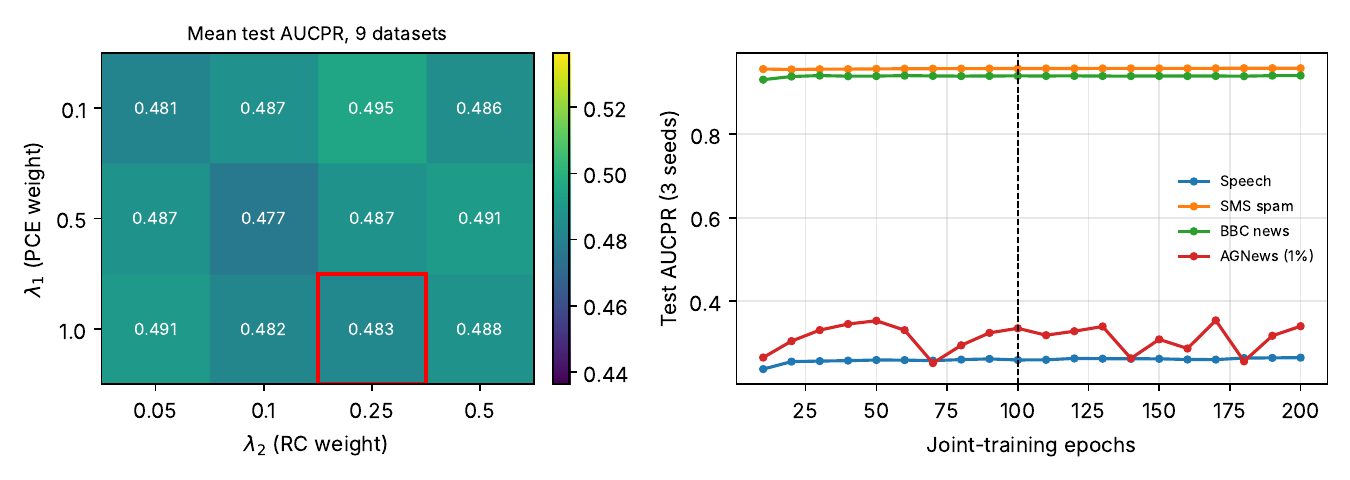}
\caption{Sensitivity. \textbf{Left:} mean test AUCPR over nine embedding datasets for each $(\lambda_1, \lambda_2)$; the colour scale spans $\pm 0.05$ around the mean so the small spread is shown at true scale; the red box is the default. \textbf{Right:} test AUCPR against the number of joint-training epochs (dashed line: default of 100).}
\label{fig:robustness}
\end{figure}

\section{Per-dataset results}
\label{app:full}
Tables~\ref{tab:highdimfull}--\ref{tab:mvtecfull} report per-dataset and per-category results for the main experiments.

\begin{table}[h]
\centering
\caption{Text and image embedding datasets at the original 5\% anomaly rate with full labels: AUCPR (mean of five seeds).}
\label{tab:highdimfull}
\small
\begin{tabular}{lccccccc}
\toprule
Dataset & TripletAD & SupCon & DeepSAD & DevNet & PReNet & CatBoost & RF \\
\midrule
20News & 0.579 & 0.586 & \textbf{0.623} & \textbf{0.623} & 0.618 & 0.586 & 0.449 \\
AGNews & 0.608 & 0.591 & \textbf{0.620} & 0.462 & 0.453 & 0.541 & 0.339 \\
Amazon & 0.324 & 0.330 & \textbf{0.346} & 0.335 & 0.320 & 0.333 & 0.161 \\
IMDB & 0.340 & 0.346 & \textbf{0.436} & 0.304 & 0.292 & 0.335 & 0.183 \\
Yelp & 0.515 & 0.503 & \textbf{0.564} & 0.488 & 0.462 & 0.499 & 0.276 \\
CIFAR-10 & 0.468 & 0.420 & 0.397 & 0.469 & \textbf{0.487} & 0.423 & 0.359 \\
FashionMNIST & 0.772 & 0.763 & \textbf{0.775} & 0.752 & 0.769 & 0.771 & 0.754 \\
SVHN & 0.554 & 0.543 & 0.515 & 0.581 & \textbf{0.593} & 0.512 & 0.401 \\
\bottomrule
\end{tabular}

\end{table}

\begin{table}[h]
\centering
\caption{Anomaly-rarity sweep with a single fixed anomaly subsample (drawn once with seed 0 and reused for all five split seeds). The conclusions match Table~\ref{tab:rarity}: \method{} has the best mean at 2\% (0.350) and 1\% (0.230), and its lead over DevNet grows from 4\% to 27\%.}
\label{tab:rarityfixed}
\small
\begin{tabular}{lcccccc}
\toprule
Dataset & TripletAD & DeepSAD & DevNet & PReNet & CatBoost & RF \\
\midrule
\multicolumn{7}{l}{\emph{Anomaly rate 1\%}} \\
\quad 20News & \textbf{0.287} & 0.218 & 0.230 & 0.249 & 0.152 & 0.103 \\
\quad AGNews & \textbf{0.201} & 0.195 & 0.128 & 0.126 & 0.103 & 0.070 \\
\quad Amazon & \textbf{0.124} & 0.114 & 0.081 & 0.074 & 0.069 & 0.021 \\
\quad IMDB & 0.175 & \textbf{0.191} & 0.104 & 0.080 & 0.067 & 0.031 \\
\quad Yelp & \textbf{0.290} & 0.254 & 0.190 & 0.168 & 0.122 & 0.043 \\
\quad CIFAR-10 & 0.113 & 0.138 & 0.143 & 0.148 & \textbf{0.190} & 0.145 \\
\quad FashionMNIST & \textbf{0.534} & 0.524 & 0.408 & 0.424 & 0.502 & 0.530 \\
\quad SVHN & 0.119 & 0.069 & \textbf{0.163} & 0.159 & 0.081 & 0.051 \\
\quad \emph{Average} & \textbf{0.230} & 0.213 & 0.181 & 0.178 & 0.161 & 0.124 \\
\midrule
\multicolumn{7}{l}{\emph{Anomaly rate 2\%}} \\
\quad 20News & 0.467 & 0.545 & 0.543 & \textbf{0.553} & 0.446 & 0.346 \\
\quad AGNews & 0.412 & \textbf{0.435} & 0.266 & 0.276 & 0.325 & 0.168 \\
\quad Amazon & \textbf{0.210} & 0.202 & 0.146 & 0.139 & 0.139 & 0.055 \\
\quad IMDB & 0.219 & \textbf{0.222} & 0.142 & 0.136 & 0.158 & 0.053 \\
\quad Yelp & 0.274 & \textbf{0.278} & \textbf{0.277} & 0.256 & 0.215 & 0.097 \\
\quad CIFAR-10 & 0.229 & 0.184 & 0.266 & \textbf{0.287} & 0.213 & 0.169 \\
\quad FashionMNIST & \textbf{0.682} & 0.680 & 0.631 & 0.625 & 0.657 & 0.633 \\
\quad SVHN & 0.310 & 0.191 & 0.305 & \textbf{0.312} & 0.208 & 0.145 \\
\quad \emph{Average} & \textbf{0.350} & 0.342 & 0.322 & 0.323 & 0.295 & 0.208 \\
\midrule
\multicolumn{7}{l}{\emph{Anomaly rate 5\%}} \\
\quad 20News & 0.579 & \textbf{0.623} & \textbf{0.623} & 0.618 & 0.586 & 0.449 \\
\quad AGNews & 0.608 & \textbf{0.620} & 0.462 & 0.453 & 0.541 & 0.339 \\
\quad Amazon & 0.324 & \textbf{0.346} & 0.335 & 0.320 & 0.333 & 0.161 \\
\quad IMDB & 0.340 & \textbf{0.436} & 0.304 & 0.292 & 0.335 & 0.183 \\
\quad Yelp & 0.515 & \textbf{0.564} & 0.488 & 0.462 & 0.499 & 0.276 \\
\quad CIFAR-10 & 0.468 & 0.397 & 0.469 & \textbf{0.487} & 0.423 & 0.359 \\
\quad FashionMNIST & 0.772 & \textbf{0.775} & 0.752 & 0.769 & 0.771 & 0.754 \\
\quad SVHN & 0.554 & 0.515 & 0.581 & \textbf{0.593} & 0.512 & 0.401 \\
\quad \emph{Average} & 0.520 & \textbf{0.535} & 0.502 & 0.499 & 0.500 & 0.365 \\
\bottomrule
\end{tabular}

\end{table}

\begin{table}[h]
\centering
\caption{Text and image benchmarks at 1\%, 2\% and their original 5\% anomaly rate: AUCPR (mean of five seeds, anomaly subsample redrawn per seed).}
\label{tab:rarityfull}
\small
\begin{tabular}{lccccccc}
\toprule
Dataset & TripletAD & SupCon & DeepSAD & DevNet & PReNet & CatBoost & RF \\
\midrule
\multicolumn{8}{l}{\emph{Anomaly rate 1\%}} \\
\quad 20News & \textbf{0.341} & 0.269 & 0.212 & \textbf{0.341} & 0.329 & 0.152 & 0.106 \\
\quad AGNews & \textbf{0.325} & 0.274 & 0.262 & 0.175 & 0.171 & 0.159 & 0.082 \\
\quad Amazon & \textbf{0.111} & 0.108 & 0.095 & 0.099 & 0.107 & 0.057 & 0.023 \\
\quad IMDB & \textbf{0.126} & 0.049 & 0.118 & 0.095 & 0.081 & 0.069 & 0.022 \\
\quad Yelp & 0.181 & 0.155 & \textbf{0.196} & 0.151 & 0.150 & 0.117 & 0.047 \\
\quad CIFAR-10 & 0.135 & 0.151 & 0.130 & 0.179 & \textbf{0.186} & 0.142 & 0.116 \\
\quad FashionMNIST & \textbf{0.575} & 0.539 & 0.573 & 0.411 & 0.431 & 0.490 & 0.491 \\
\quad SVHN & 0.154 & 0.174 & 0.092 & 0.183 & \textbf{0.203} & 0.089 & 0.061 \\
\quad \emph{Average} & \textbf{0.244} & 0.215 & 0.210 & 0.204 & 0.207 & 0.159 & 0.118 \\
\midrule
\multicolumn{8}{l}{\emph{Anomaly rate 2\%}} \\
\quad 20News & 0.441 & \textbf{0.476} & 0.396 & 0.463 & 0.447 & 0.367 & 0.250 \\
\quad AGNews & 0.332 & 0.340 & \textbf{0.392} & 0.304 & 0.300 & 0.293 & 0.169 \\
\quad Amazon & \textbf{0.240} & 0.161 & 0.169 & 0.158 & 0.152 & 0.142 & 0.048 \\
\quad IMDB & \textbf{0.197} & 0.185 & 0.182 & 0.168 & 0.161 & 0.136 & 0.061 \\
\quad Yelp & 0.319 & 0.285 & \textbf{0.364} & 0.294 & 0.288 & 0.261 & 0.099 \\
\quad CIFAR-10 & 0.281 & 0.248 & 0.217 & \textbf{0.295} & 0.287 & 0.250 & 0.191 \\
\quad FashionMNIST & \textbf{0.660} & 0.653 & 0.621 & 0.603 & 0.600 & 0.612 & 0.575 \\
\quad SVHN & 0.295 & 0.284 & 0.266 & \textbf{0.376} & 0.353 & 0.247 & 0.173 \\
\quad \emph{Average} & \textbf{0.346} & 0.329 & 0.326 & 0.333 & 0.324 & 0.289 & 0.196 \\
\midrule
\multicolumn{8}{l}{\emph{Anomaly rate 5\%}} \\
\quad 20News & 0.579 & 0.586 & \textbf{0.623} & \textbf{0.623} & 0.618 & 0.586 & 0.449 \\
\quad AGNews & 0.608 & 0.591 & \textbf{0.620} & 0.462 & 0.453 & 0.541 & 0.339 \\
\quad Amazon & 0.324 & 0.330 & \textbf{0.346} & 0.335 & 0.320 & 0.333 & 0.161 \\
\quad IMDB & 0.340 & 0.346 & \textbf{0.436} & 0.304 & 0.292 & 0.335 & 0.183 \\
\quad Yelp & 0.515 & 0.503 & \textbf{0.564} & 0.488 & 0.462 & 0.499 & 0.276 \\
\quad CIFAR-10 & 0.468 & 0.420 & 0.397 & 0.469 & \textbf{0.487} & 0.423 & 0.359 \\
\quad FashionMNIST & 0.772 & 0.763 & \textbf{0.775} & 0.752 & 0.769 & 0.771 & 0.754 \\
\quad SVHN & 0.554 & 0.543 & 0.515 & 0.581 & \textbf{0.593} & 0.512 & 0.401 \\
\quad \emph{Average} & 0.520 & 0.510 & \textbf{0.535} & 0.502 & 0.499 & 0.500 & 0.365 \\
\bottomrule
\end{tabular}

\end{table}

\begin{table}[h]
\centering
\caption{Eight text and image embedding datasets at their original 5\% anomaly rate, with fewer anomalies in training: mean AUCPR. Rows give the range of training anomalies over datasets, with their share of the training data in brackets. Cells with fewer than ten training anomalies are skipped, so the last row covers five datasets.}
\label{tab:sparse}
\small
\begin{tabular}{lccccccc}
\toprule
Training anomalies (share) & TripletAD & SupCon & DeepSAD & DevNet & PReNet & CatBoost & RF \\
\midrule
27--87 (1.28--1.30\%), 8 sets & 0.445 & 0.389 & 0.402 & \textbf{0.448} & 0.439 & 0.369 & 0.258 \\
10--35 (0.48--0.52\%), 8 sets & 0.357 & 0.263 & 0.277 & \textbf{0.358} & 0.353 & 0.249 & 0.194 \\
11--17 (0.25--0.26\%), 5 sets & 0.319 & 0.228 & 0.240 & \textbf{0.334} & 0.326 & 0.232 & 0.183 \\
\midrule
TripletAD wins & -- & 21/21 & 20/21 & 9/21 & 10/21 & 20/21 & 21/21 \\
\bottomrule
\end{tabular}

\end{table}

\begin{table}[h]
\centering
\caption{Raw tabular datasets with full labels: AUCPR (mean of five seeds). PReNet and DeepSAD were not run on Backdoor, Cover, Fraud and HTTP.}
\label{tab:tabular}
\small
\begin{tabular}{lcccccc}
\toprule
Dataset & TripletAD & DeepSAD & DevNet & PReNet & CatBoost & RF \\
\midrule
Backdoor & 0.974 & -- & 0.882 & -- & \textbf{0.998} & 0.996 \\
Cover & 0.995 & -- & 0.956 & -- & \textbf{1.000} & 0.999 \\
Fraud & 0.773 & -- & 0.743 & -- & \textbf{0.866} & 0.849 \\
HTTP & 0.998 & -- & 0.997 & -- & \textbf{0.999} & \textbf{0.999} \\
Waveform & 0.505 & 0.538 & 0.209 & 0.234 & \textbf{0.605} & 0.550 \\
Yeast & 0.588 & 0.536 & 0.487 & 0.490 & \textbf{0.662} & 0.655 \\
ALOI & 0.205 & 0.145 & 0.046 & 0.048 & 0.420 & \textbf{0.538} \\
Campaign & 0.450 & 0.586 & 0.503 & 0.516 & \textbf{0.662} & 0.636 \\
CelebA & 0.250 & 0.435 & 0.337 & 0.323 & \textbf{0.466} & 0.390 \\
Census & 0.415 & 0.603 & 0.539 & 0.529 & \textbf{0.697} & 0.619 \\
\bottomrule
\end{tabular}

\end{table}

\begin{table}[h]
\centering
\caption{Text and image embedding datasets at their original 5\% anomaly rate with fewer anomalies in training: AUCPR (mean of five seeds), with the number of training anomalies and their share of the training data.}
\label{tab:sparsefull}
\small
\setlength{\tabcolsep}{4pt}
\begin{tabular}{llccccccc}
\toprule
Dataset & Training anomalies & TripletAD & SupCon & DeepSAD & DevNet & PReNet & CatBoost & RF \\
\midrule
20News & 27 (1.30\%) & 0.468 & 0.400 & 0.416 & \textbf{0.544} & 0.511 & 0.300 & 0.264 \\
AGNews & 87 (1.29\%) & 0.523 & 0.479 & \textbf{0.527} & 0.463 & 0.445 & 0.447 & 0.223 \\
Amazon & 87 (1.29\%) & 0.310 & 0.220 & 0.244 & \textbf{0.314} & 0.303 & 0.245 & 0.109 \\
IMDB & 87 (1.29\%) & \textbf{0.309} & 0.265 & 0.305 & 0.280 & 0.259 & 0.245 & 0.113 \\
Yelp & 87 (1.29\%) & \textbf{0.486} & 0.412 & 0.454 & 0.468 & 0.448 & 0.376 & 0.185 \\
CIFAR-10 & 46 (1.30\%) & 0.318 & 0.263 & 0.254 & 0.343 & \textbf{0.362} & 0.319 & 0.247 \\
FashionMNIST & 55 (1.29\%) & \textbf{0.719} & 0.672 & 0.704 & 0.687 & 0.698 & 0.676 & 0.656 \\
SVHN & 45 (1.28\%) & 0.430 & 0.400 & 0.314 & \textbf{0.486} & 0.483 & 0.345 & 0.264 \\
20News & 10 (0.48\%) & \textbf{0.379} & 0.220 & 0.197 & 0.358 & 0.354 & 0.179 & 0.166 \\
AGNews & 35 (0.52\%) & \textbf{0.446} & 0.283 & 0.359 & 0.348 & 0.325 & 0.256 & 0.145 \\
Amazon & 35 (0.52\%) & 0.242 & 0.150 & 0.148 & \textbf{0.258} & 0.246 & 0.145 & 0.078 \\
IMDB & 35 (0.52\%) & \textbf{0.221} & 0.152 & 0.192 & 0.214 & 0.203 & 0.164 & 0.075 \\
Yelp & 35 (0.52\%) & 0.396 & 0.317 & 0.318 & \textbf{0.430} & 0.416 & 0.234 & 0.148 \\
CIFAR-10 & 18 (0.51\%) & 0.188 & 0.162 & 0.184 & 0.223 & \textbf{0.225} & 0.165 & 0.158 \\
FashionMNIST & 22 (0.52\%) & \textbf{0.637} & 0.547 & 0.611 & 0.627 & 0.632 & 0.600 & 0.583 \\
SVHN & 18 (0.52\%) & 0.348 & 0.278 & 0.204 & 0.409 & \textbf{0.424} & 0.249 & 0.200 \\
AGNews & 17 (0.25\%) & 0.302 & 0.221 & 0.193 & \textbf{0.320} & 0.311 & 0.210 & 0.126 \\
Amazon & 17 (0.25\%) & 0.184 & 0.114 & 0.097 & \textbf{0.207} & 0.196 & 0.123 & 0.068 \\
IMDB & 17 (0.25\%) & \textbf{0.202} & 0.120 & 0.089 & 0.197 & 0.187 & 0.124 & 0.063 \\
Yelp & 17 (0.25\%) & 0.300 & 0.213 & 0.226 & \textbf{0.325} & 0.310 & 0.165 & 0.098 \\
FashionMNIST & 11 (0.26\%) & 0.608 & 0.471 & 0.597 & 0.624 & \textbf{0.628} & 0.540 & 0.562 \\
\bottomrule
\end{tabular}

\end{table}

\begin{table}[h]
\centering
\caption{MVTec-AD per category: AUCPR (mean of five seeds), with the number of training defects and their share of the training data. ``--'': DeepSAD's data loader failed.}
\label{tab:mvtecfull}
\scriptsize
\begin{tabular}{llccccccc}
\toprule
Category & Training defects & TripletAD & SupCon & DeepSAD & DevNet & PReNet & CatBoost & RF \\
\midrule
bottle & 44 (21.6\%) & 0.947 & \textbf{0.960} & 0.948 & 0.951 & 0.930 & 0.958 & 0.947 \\
 & 33 (17.1\%) & 0.958 & \textbf{0.959} & 0.948 & 0.924 & 0.913 & \textbf{0.958} & 0.955 \\
 & 22 (12.1\%) & \textbf{0.963} & 0.936 & 0.953 & 0.927 & 0.942 & 0.954 & 0.949 \\
 & 11 (6.4\%) & 0.947 & 0.926 & \textbf{0.948} & 0.920 & 0.920 & \textbf{0.948} & \textbf{0.948} \\
\addlinespace[2pt]
cable & 64 (24.5\%) & 0.534 & 0.562 & 0.545 & 0.540 & 0.555 & 0.543 & \textbf{0.571} \\
 & 48 (19.6\%) & \textbf{0.594} & 0.544 & 0.514 & 0.537 & 0.547 & 0.570 & 0.591 \\
 & 32 (14.0\%) & \textbf{0.582} & 0.526 & 0.492 & 0.497 & 0.488 & 0.526 & 0.539 \\
 & 16 (7.5\%) & \textbf{0.568} & 0.504 & 0.495 & 0.452 & 0.455 & 0.548 & 0.562 \\
\addlinespace[2pt]
capsule & 76 (31.0\%) & \textbf{0.743} & 0.723 & 0.704 & 0.661 & 0.659 & 0.675 & 0.669 \\
 & 57 (25.2\%) & \textbf{0.725} & 0.709 & 0.704 & 0.659 & 0.645 & 0.677 & 0.690 \\
 & 38 (18.4\%) & \textbf{0.724} & 0.685 & 0.676 & 0.628 & 0.629 & 0.698 & 0.671 \\
 & 19 (10.1\%) & 0.708 & 0.607 & 0.642 & 0.562 & 0.608 & \textbf{0.722} & 0.705 \\
\addlinespace[2pt]
carpet & 62 (22.4\%) & 0.662 & 0.687 & 0.639 & 0.735 & 0.708 & \textbf{0.737} & 0.698 \\
 & 46 (17.6\%) & 0.666 & 0.692 & 0.640 & \textbf{0.734} & 0.696 & 0.728 & 0.703 \\
 & 31 (12.6\%) & 0.651 & 0.664 & 0.602 & 0.711 & 0.675 & \textbf{0.714} & 0.687 \\
 & 15 (6.5\%) & 0.644 & 0.686 & 0.575 & \textbf{0.710} & 0.687 & 0.705 & 0.667 \\
\addlinespace[2pt]
grid & 40 (16.7\%) & \textbf{0.718} & 0.696 & 0.550 & 0.708 & 0.673 & 0.687 & 0.678 \\
 & 30 (13.1\%) & \textbf{0.699} & 0.674 & 0.549 & 0.698 & 0.683 & 0.671 & 0.645 \\
 & 20 (9.1\%) & 0.659 & 0.647 & 0.522 & \textbf{0.687} & 0.655 & 0.648 & 0.624 \\
 & 10 (4.8\%) & 0.581 & \textbf{0.593} & 0.496 & 0.552 & 0.570 & 0.565 & 0.546 \\
\addlinespace[2pt]
hazelnut & 49 (14.0\%) & 0.637 & 0.718 & 0.641 & 0.689 & 0.686 & \textbf{0.781} & 0.752 \\
 & 36 (10.7\%) & 0.649 & 0.733 & 0.642 & 0.684 & 0.672 & \textbf{0.774} & 0.730 \\
 & 24 (7.4\%) & 0.640 & 0.719 & 0.615 & 0.638 & 0.640 & \textbf{0.746} & 0.729 \\
 & 12 (3.8\%) & 0.620 & 0.670 & 0.553 & 0.589 & 0.640 & \textbf{0.715} & 0.683 \\
\addlinespace[2pt]
leather & 64 (24.8\%) & 0.970 & 0.955 & 0.938 & \textbf{0.995} & 0.951 & 0.992 & 0.992 \\
 & 48 (19.8\%) & 0.935 & 0.949 & 0.950 & \textbf{0.994} & 0.966 & 0.992 & 0.991 \\
 & 32 (14.2\%) & 0.920 & 0.956 & 0.943 & \textbf{0.993} & 0.973 & 0.992 & 0.990 \\
 & 16 (7.6\%) & 0.945 & 0.969 & 0.936 & \textbf{0.992} & 0.991 & 0.990 & 0.985 \\
\addlinespace[2pt]
metal nut & 65 (27.8\%) & \textbf{0.766} & 0.732 & 0.578 & 0.752 & 0.747 & 0.726 & 0.681 \\
 & 48 (22.1\%) & \textbf{0.772} & 0.720 & 0.580 & 0.751 & 0.738 & 0.713 & 0.682 \\
 & 32 (15.9\%) & 0.716 & \textbf{0.731} & 0.560 & 0.716 & 0.727 & 0.668 & 0.631 \\
 & 16 (8.6\%) & \textbf{0.712} & 0.656 & 0.498 & 0.688 & 0.705 & 0.607 & 0.592 \\
\addlinespace[2pt]
pill & 98 (32.3\%) & 0.597 & 0.620 & 0.584 & \textbf{0.637} & 0.628 & 0.601 & 0.578 \\
 & 73 (26.3\%) & 0.632 & 0.602 & 0.580 & \textbf{0.636} & 0.617 & 0.603 & 0.592 \\
 & 49 (19.3\%) & \textbf{0.624} & 0.601 & 0.547 & 0.611 & 0.611 & 0.610 & 0.604 \\
 & 24 (10.5\%) & 0.556 & 0.542 & 0.527 & \textbf{0.593} & 0.586 & 0.584 & 0.570 \\
\addlinespace[2pt]
screw & 83 (24.7\%) & 0.598 & 0.597 & 0.493 & \textbf{0.602} & 0.594 & 0.559 & 0.512 \\
 & 62 (19.7\%) & 0.597 & 0.601 & 0.496 & 0.602 & \textbf{0.604} & 0.519 & 0.487 \\
 & 41 (13.9\%) & \textbf{0.587} & 0.545 & 0.465 & 0.554 & 0.552 & 0.468 & 0.472 \\
 & 20 (7.3\%) & \textbf{0.472} & 0.468 & 0.424 & 0.457 & 0.464 & 0.387 & 0.388 \\
\addlinespace[2pt]
tile & 59 (24.4\%) & 0.786 & \textbf{0.822} & 0.733 & 0.796 & 0.798 & 0.790 & 0.783 \\
 & 44 (19.4\%) & 0.812 & \textbf{0.816} & 0.751 & 0.794 & 0.790 & 0.785 & 0.797 \\
 & 29 (13.7\%) & 0.782 & \textbf{0.805} & 0.737 & 0.766 & 0.774 & 0.773 & 0.790 \\
 & 14 (7.1\%) & 0.731 & 0.679 & 0.692 & 0.653 & 0.679 & 0.751 & \textbf{0.769} \\
\addlinespace[2pt]
toothbrush & 21 (29.6\%) & 0.892 & \textbf{0.900} & -- & 0.890 & 0.891 & 0.880 & 0.879 \\
 & 15 (23.1\%) & 0.842 & \textbf{0.851} & -- & 0.792 & 0.809 & 0.834 & 0.832 \\
 & 10 (16.7\%) & 0.823 & 0.762 & -- & 0.817 & 0.805 & \textbf{0.836} & 0.821 \\
\addlinespace[2pt]
transistor & 28 (12.8\%) & 0.586 & 0.616 & 0.538 & 0.597 & 0.587 & \textbf{0.716} & 0.698 \\
 & 21 (9.9\%) & 0.535 & 0.554 & 0.498 & 0.543 & 0.571 & \textbf{0.683} & 0.623 \\
 & 14 (6.8\%) & 0.614 & 0.583 & 0.469 & 0.542 & 0.571 & \textbf{0.667} & 0.640 \\
\addlinespace[2pt]
wood & 42 (18.4\%) & 0.784 & 0.789 & 0.692 & 0.716 & 0.694 & 0.781 & \textbf{0.796} \\
 & 31 (14.3\%) & 0.752 & 0.767 & 0.673 & 0.725 & 0.690 & 0.765 & \textbf{0.772} \\
 & 21 (10.1\%) & 0.756 & 0.718 & 0.666 & 0.664 & 0.658 & \textbf{0.772} & 0.761 \\
 & 10 (5.1\%) & \textbf{0.726} & 0.617 & 0.622 & 0.620 & 0.636 & 0.664 & 0.707 \\
\addlinespace[2pt]
zipper & 83 (30.4\%) & 0.952 & 0.948 & 0.883 & \textbf{0.966} & 0.952 & 0.940 & 0.898 \\
 & 62 (24.6\%) & \textbf{0.966} & 0.951 & 0.785 & 0.959 & 0.953 & 0.939 & 0.900 \\
 & 41 (17.7\%) & \textbf{0.964} & 0.935 & 0.767 & 0.956 & 0.945 & 0.909 & 0.873 \\
 & 20 (9.5\%) & \textbf{0.937} & 0.915 & 0.721 & 0.931 & 0.933 & 0.880 & 0.854 \\
\bottomrule
\end{tabular}

\end{table}

\end{document}